# A Multimodal Eye Movement Dataset and a Multimodal Eye Movement Segmentation Analysis

WOLFGANG FUHL and ENKELEJDA KASNECI, University Tübingen, Germany

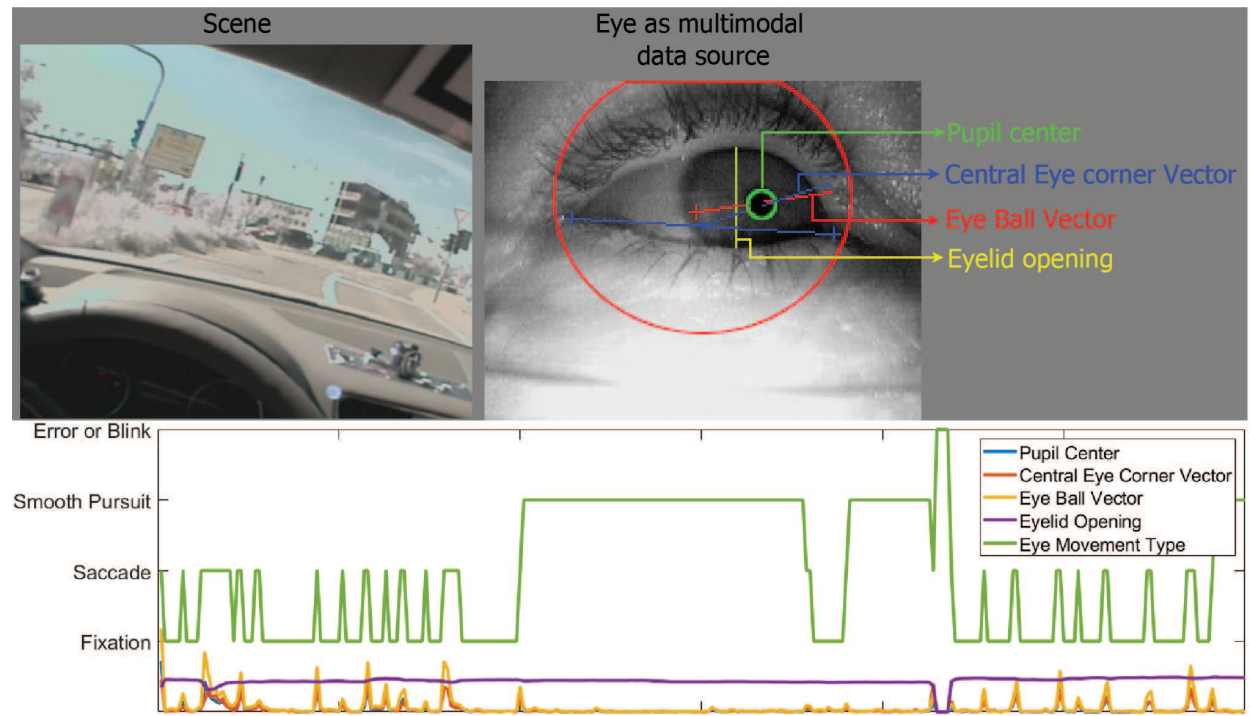

Fig. 1. data streams which can be extracted from a human eye image and used for eye movement classification.

We present a new dataset with annotated eye movements. The dataset consists of over 800,000 gaze points recorded during a car ride in the real world and in the simulator. In total, the eye movements of 19 subjects were annotated. In this dataset there are several data sources such as the eyelid closure, the pupil center, the optical vector, and a vector into the pupil center starting from the center of the eye corners. These different data sources are analyzed and evaluated individually as well as in combination with respect to their goodness of fit for eye movement classification. These results will help developers of real-time systems and algorithms to find the best data sources for their application. Also, new algorithms can be trained and evaluated on this data set. The data and the Matlab code can be downloaded here https://atreus.informatik.uni-tuebingen.de/seafile/d/8e2ab8c3fdd444e1a135/?p=%2FA%20Multimodal%20Eye%20Movement%20Dataset%20and%20...&mode=list.

CCS Concepts: • **Computing methodologies** → **Neural networks**; **Classification and regression trees**; • **Human-centered computing** → **Empirical studies in ubiquitous and mobile computing**;

Additional Key Words and Phrases: Eye Movements, Data set, Classification, Driving, Real World, Machine Learning, Segmentation

**ACM Reference Format:**
Wolfgang Fuhl and Enkelejda Kasneci. 2018. A Multimodal Eye Movement Dataset and a Multimodal Eye Movement Segmentation Analysis. In . ACM, New York, NY, USA, 12 pages. https://doi.org/10.1145/1122445.1122456

## 1 INTRODUCTION

The eyes are an increasingly important source of information [10]. Current research in eye movements is concerned with cognitive states [64], workload of people [65], attention assurance in autonomous driving vehicles [54], and gaze forecasting [61]. In addition, classical research in the eye domain, such as feature extraction [20, 23, 24, 34, 36, 43–46, 49, 50, 60], eye movement classification [1, 14, 16, 25, 37, 47, 59], and gaze point determination [21, 56], is far from complete.

The application fields of an eye tracker are wide-ranging and include expertise determination [4], human computer interaction [11], human robot interaction [3], improved remote assistance [71], visualizations [33, 35, 55], facilitating the work of surgeons [2, 5, 7, 8, 43], and much more. Due to this variety of possible applications, eye trackers must perform reliably under a wide range of conditions, which creates a great many challenges in image processing [10] but also in eye movement classification [37]. Many Classical eye movement classification algorithms use a variety of thresholds which are applied to the data [1, 52]. More modern algorithms use a wide variety of machine learning techniques [13, 28–31] to do this [15, 17, 37, 59, 77] but it is still very challenging. The modern methods have the advantage that the algorithms can be adapted to different eye trackers through training and annotated data. The disadvantage this creates is the need for a large amount of annotated data with a high quality. Some works [6, 26, 37, 48] have therefore dealt with the generation and simulation of eye movements.

In this work, we present a new dataset that includes annotations for fixations, saccades, and smooth pursuits. Due to the driving context it can also be used for scan path analysis [12, 18] and saliency prediction [32, 51]. Currently, this is the world's largest dataset along with new metrics such as the optical vector, relative eye opening state, and a vector computed from the pupil center and the center of the eye corners. Since this dataset is based on the image data of [21, 39], the segmentations for the pupil as well as the sclera [19, 22, 39] are also given as well as the eyeball parameters and the optical vector. Also, this work includes an evaluation of different machine learning algorithms to assess the goodness of individual features. The contribution of this work is listed below as a bullet point list.

(1) The first contribution of this work is an eyeball annotated dataset which has already been annotated with semantic segmentations, the eyeball, and the optical vector by previous work [21, 39]. These data come from the car driving context and was recorded in a simulator as well as during real driving.
(2) The presented dataset also contains more extracted features than previous datasets regarding eye movements. As features we have the optical vector, the pupil center, a vector which has its origin exactly between the eye corners, and the degree of aperture of the eyelid [40–42]. Also, the movements in the x,y and depth dimension are additionally provided.
(3) With this multitude of features, we perform an analysis that highlights the importance and contribution of each feature. To the best of our knowledge, this is also the first work that looks at these features and their combination. This allows algorithm developers to more easily select the data they need.
(4) To our knowledge, the presented dataset is also the world's largest dataset of annotated eye movements. This includes fixations, saccades, smooth pursuits, and errors in the data. The recordings come from long-term recordings in a driving simulator and from real-world car driving.

## 2 RELATED WORK

The two best known algorithms are IDT and IVT (Identification by Dispersion Threshold, Identification by Velocity Threshold) [69]. Here, different thresholds are used to limit the dispersion of the data points and the length of the segments. For IVT, only velocity is considered and a threshold is used to distinguish between fiaxtion and saccade. For IVT there is also an approach which adaptively determines the threshold for velocity [9]. For filtering and smoothing the signal, a Kalman filter has also been presented [58]. Since the Kalman filter makes predictions, this smoothing can be used online. Also IVT was extended by a threshold for the segment length [57]. An approach which uses the $\chi^2$-test for smoothing was published in [57]. An extension of the IDT algorithm was presented using the F-tests scatter [74]. Here, the F-test decides the class thereby bypassing a fixed threshold. Since the F-test is very susceptible to noise, covariance was used instead of the F-test in [75]. However, the covariance approach has the disadvantage that three thresholds are now needed. In [57], a minimal spanning tree was computed to group the data into clusters. These clusters correspond to eye movement types. Since this algorithm requires all the data to compute, it cannot be used online.

In the field of machine learning, the first approaches were used to compute the threshold via statistics. The first approach used Hidden Markov Models (HMM) [57] and applied them to velocity. The model itself has two states and distinguishes between fixations and saccades. The first extension of this approach was presented in [72]. Here, automatic parameter determination was introduced. More recent approaches deal with newer eye movement types such as smooth pursuits and post saccadic movements (PSM). The first algorithm for PSM detection was presented in [67]. In the following year an algorithm which considers both eyes was presented [73]. The latter two algorithms use adaptive thresholding, the second assuming that both eyes perform the same movement. For four eye movements, an approach using adaptive thresholds was described in [63]. This can only be used offline, since it also includes some preprocessing steps. For high sampling rates, this algorithm has been extended [62]. The novelty in this approach is that a coarse segmentation is created after preprocessing. This segmentation is then further refined until the entire data is annotated.

Recent approaches to eye movement classification use deep neural networks (DNN) or random forrests (RF). The first approach from this field is published in [53]. This approach transforms the data points in a fixed window into frequency space and then uses a DNN for classification. In [77], an approach using RF was presented. The algorithm was trained to work with different sampling rates. This was achieved by preprocessing the data using cubic spline interpolating and thus mapping it to a fixed sampling rate. Fourteen statistical features were also computed, which are used for classification. A rule-based approach, which can be fed with different data, was published in [17]. It learns rules consisting of threshold values and segments the individual data streams. Each segment combination is assigned a class. For the use of arbitrary machine learning algorithms the feature extractor histogram of oriented velocities (HOV) was presented [15]. Here, the HOVs are computed on the data and can then be used with any machine learning method. GazeNet [76] is another approach which uses deep neural networks together with LSTM (Long Short Term Memory) cells. Semantic eye movement classification in combination with fully convolutional networks was presented in [37]. Here variational autoencoders were used to generate data too. In [59] a new dataset as well as an algorithm based on RF was presented.

## 3 ANNOTATION PROCESS

For the annotation, we summarized and normalized the data from [21, 39]. In the first step, the optical vectors was normalized to a magnitude of one. The pupil center was normalized to the range of values of the image width and height, so that the new x and y coordinate multiplied by the resolution gives the coordinates in pixels. The center was also calculated from the corners of the eyes and from this the vector to the center of the pupil was calculated. This vector was also normalized to the image width and height, but with the difference that it can contain both negative and positive values. To calculate the pupil center in pixels from this vector between the pupil and the eye corner center, the vector must be multiplied by the resolution and added to the eye corner center.

From all these values, the difference between two points was calculated as well as the difference for each axis and stored in a matrix. Values in which the detection failed based on the work of [39] were marked as errors. Also, all values using an error entry in the calculation were marked as errors. Subsequently, fixations and saccades were annotated over physiologically determined thresholds [52]. These segments were manually inspected and the ranges were adjusted. In this step, fixations were also reclassified to Smooth Purisuites. Finally, the approach of [37] was used to make the annotation as consistent as possible together with different validation procedures [27, 38]. Here, the approach was trained on 50% of a person's data at a time and then applied to the remaining 50%. This process was repeated several times. Subsequently, all areas in which there was a difference between the annotation and the automated detection were inspected manually.

## 4 DATASET DESCRIPTION

- F1: Normalized euclidean pupil center distance between two frames.
- F2: Normalized euclidean eyelid center vector distance between two frames.
- F3: Normalized euclidean optical vector distance between two frames.
- F4: Eyelid opening in relation to eye width (eye corner distance).
- F5: Normalized pupil center distance in x direction.
- F6: Normalized pupil center distance in y direction.
- F7: Normalized eyelid center vector distance in x direction.
- F8: Normalized eyelid center vector distance in y direction.
- F9: Normalized optical vector distance in x direction.
- F10: Normalized optical vector distance in y direction.
- F11: Normalized optical vector distance in z direction.

The presented data set includes eleven features which are listed and described in the List 4. The image data comes from [21, 39]. The normalization of the data is described in the section 3. As can be seen in the List 4, these eleven features are four main features (F1-F4) and their differences into different dimensions (F5-F11). In total, our dataset has 866,050 annotated entries of which 154,242 are eye movement types segments and 10,580 are error segments. In total, 3.89% of the dataset is flagged as error based on the entries. More detailed information on all data types can be found in the Table 1. Our dataset includes car driving recordings of 19 people in total with the first 10 having driven in a simulator and the last 9 having performed real car rides. We also take this split for training (driving simulator) and testing (real driving) in the 5 section, where the multimodal data analysis is described. The eye tracker used was a Dikablis Pro with 25 frames per second. This low sampling rate makes it even harder to distinguish the different eye movement types.

Our data is provided in a Matlab data container (Mat file). Here the division into subjects as well as the division of the segments per eye movement type with amount of entries, start index, and stop index is given. Under Features you can find all features with the corresponding label.

Table 1 shows all statistics of our data set. Due to the low sampling rate of the eye tracker, it is also possible that there are several consecutive saccades with no fixation in between. This is due to the fact that the fixation occurred between two images. However, these annotations occur very rarely, which can also be seen in the statistics regarding saccades in Table 1, as the mean and

Multimodal Eye Movement Dataset and Segmentation Analysis , ,

Table 1. Statistical characteristics of our dataset for each event type separately.

| Eye Movement Type | Data Points | Event Count | Mean Length | Deviation Length |
| --- | --- | --- | --- | --- |
| Fixation | 510.814 | 71.868 | 7,24 | 4,45 |
| Saccade | 112.519 | 76.183 | 1,45 | 1,05 |
| Smooth Pursuit | 208.962 | 6.191 | 32,94 | 16,75 |
| Error or Blink | 33.755 | 10.580 | 3,12 | 2,87 |

standard deviation are close to one for saccades. The longest segments are the smooth pursuits, which have a mean length of 32.94 data points as well as a standard deviation of 16.75 data points. Since these are subsequent movements, this is quite normal. The error segments in our dataset are relatively small on average, but there are also longer segments in person 15, which will be described later together with Figure 3. The fixations in our data set are up to one second long whereas these long fixations occur very rarely. Looking at the mean and standard deviation for fixations in Table 1, they are in the range of 290 milliseconds with a deviation of 180 milliseconds.

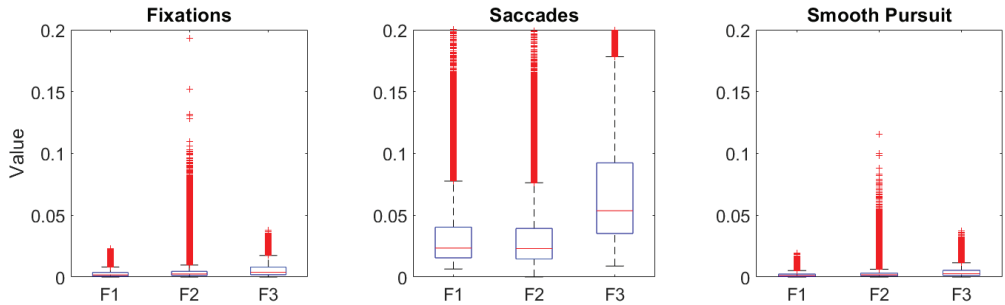

Fig. 2. Wisker plots for the feature values from F1, F2, and F3. Red crosses correspond to values that are considered as ouliers. The blue boxes represents the 75% confidence interval and the red line is the median.

Figure 2 shows the distribution of the value ranges of the main features F1, F2, and F3. The red crosses are considered outliers. The blue boxes are the 75% confidence intervals and the red line in the confidence interval is the median. Comparing the central (saccades) with the left (fixations) and right (smooth pursuit) plot, it is clear that the range of values regarding all features for saccades is significantly higher than the range of values for fixations and smooth pursuits. When looking at the left and right plot, it is noticeable that the second feature (F2) has significantly more outlier than the features F1 and F3. Since this is the vector between the eye corner center and the pupil, it can be assumed that the classification results in section 5 are significantly worse with this feature alone in comparison to F1 or F3. Another peculiarity of feature F3 is that it has a much higher range of values for saccades than features F1 and F2. F3 is the optical vector, which is also used for shift invariant calibrations. If the left plot is compared with the right plot, it is noticeable that the smooth pursuits have a minimally smaller range of values than the fixations. This is due to the fact that the fixations have significantly more transitions to and from saccades in relation to the data points, since smooth pursuits are generally much longer than fixations (See Table 1).

Figure 3 shows a section of the data from exactly 2,500 samples for each individual. As described in the legend at the very top, the color green stands for fixations, yellow for saccades, blue early smooth pursuits, and red for errors or blinks in the data. What can be seen first is that the smooth purisuites occur significantly more often in the simulator data (P1-P10). Likewise, one can see, for example, in person twelve (P12), that there can also be long saccades. As already mentioned in the statistics for table 1, this is due to successive saccades in which a very short fixation or blink occurs between two frames. In person fifteen (P15), large error segments can be seen. On the one hand, this is due to errors of the camera, which produced only distorted or black eye images in certain time ranges. Another reason for the large error segments are very strong reflections on the person's glasses. These reflections cover the complete eye and thus make the detection of the eyelids and the pupil impossible.

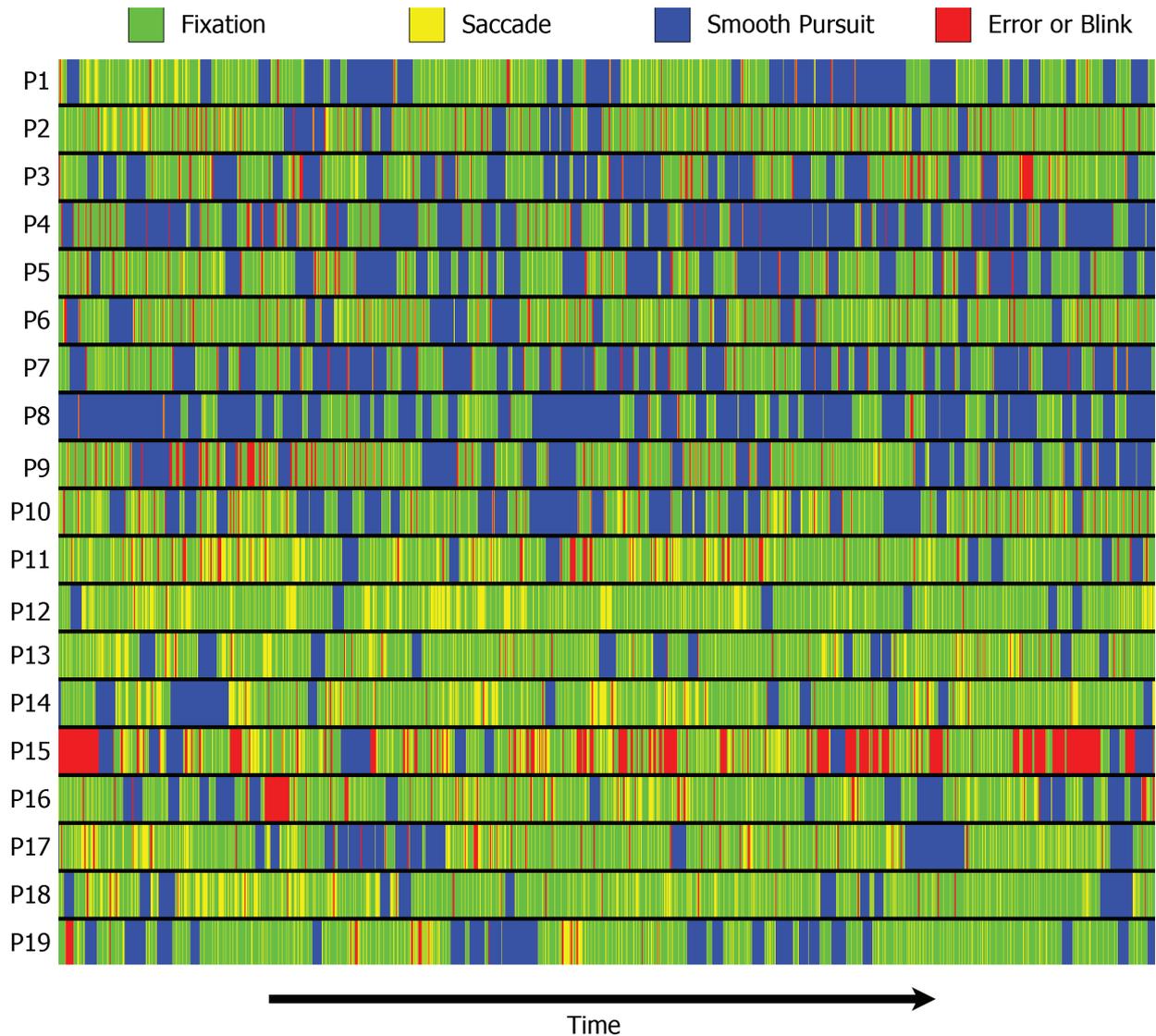

Fig. 3. Shows a small section of the data for each person. Here, the colors red represent an error, blue a smooth pursuit, green a fixation, and yellow a saccade. The clipping corresponds to exactly 2.500 samples. Larger red areas, especially for person 15, come from eye camera malfunctions or reflections that cover the entire eye.

## 5 MULTIMODAL EYE MOVEMENT CLASSIFICATION

For the evaluation of different data sources, we have chosen the two most popular machine learning methods. This would be the random forrests (RF) on the one hand and the neural networks (NN) on the other hand. For the RF we evaluated two optimization methods, this would be bagging [68] where several RFs are trained and in the end a majority decision decides the class. The other optimization method is RUS boosting [70], as it is particularly well suited for non-equilibrium datasets. Here, multiple RFs are also trained but each RF further optimizes the result of the last RF. For both approaches we always used 100 RFs and the default parameters of MatLab. This can also be seen in the attached script (Supplementary Material). For the NN we used the optimization method scaled conjugated gradients [66]. This allows very small meshes to be trained on the entire training set. The NN always had only one hidden layer with 50 neurons. As with the RF, we used the standard MatLab parameters for the NN. This can also be seen in the attached script in the supplementary material.

Table 2. In the area of machine learning approaches, we have chosen decision trees, and a tiny neural networks. For the decision trees, we evaluate the boosting and bagging training algorithms. The input data are the normalized features specified on the left side of the table. Here we evaluated different window length and used the classifiers in an **online fashion** meaning that only previous data points can be used in the window for classification of the current data point. *F=Fixation,S=Saccade,P=Smooth Pursuit,E=Error*

| Window | Features | Bagged RF | | | | Boosted RF | | | | Neural Network | | | |
|---|---|---|---|---|---|---|---|---|---|---|---|---|---|
| | Unit: | mean Intersection over Union (mIoU) * 100 | | | | | | | | | | | |
| | | F | S | P | E | F | S | P | E | F | S | P | E |
| 1 | F1,F5,F6 | 66 | 88 | 12 | 97 | 0 | 87 | 22 | 97 | 75 | 81 | 0 | 98 |
| | F2,F7,F8 | 61 | 67 | 12 | 97 | 64 | 0 | 0 | 98 | 71 | 58 | 0 | 98 |
| | F3,F9,F10,F11 | 68 | 84 | 9 | 98 | 64 | 0 | 0 | 97 | 75 | 87 | 0 | 98 |
| | F1,F5,F6,F4 | 70 | 87 | 9 | 93 | 0 | 87 | 23 | 87 | 75 | 85 | 0 | 95 |
| | F2,F7,F8,F4 | 65 | 67 | 9 | 92 | 0 | 0 | 18 | 87 | 71 | 58 | 0 | 98 |
| | F3,F9,F10,F11,F4 | 70 | 85 | 8 | 93 | 0 | 0 | 18 | 87 | 75 | 87 | 0 | 98 |
| | F1,F5,F6,F2,F7,F8 | 74 | 96 | 7 | 98 | 0 | 87 | 22 | 97 | 76 | 89 | 0 | 97 |
| | F1,F5,F6,F3,F9,F10,F11 | 73 | 88 | 7 | 98 | 0 | 87 | 22 | 98 | 75 | 87 | 0 | 98 |
| | F2,F7,F8,F3,F9,F10,F11 | 74 | 94 | 6 | 98 | 64 | 0 | 0 | 97 | 75 | 89 | 0 | 98 |
| | F1,F5,F6,F2,F7,F8,F4 | 75 | 96 | 6 | 93 | 0 | 87 | 23 | 87 | 75 | 77 | 0 | 98 |
| | F1,F5,F6,F3,F9,F10,F11,F4 | 73 | 88 | 6 | 94 | 0 | 87 | 23 | 87 | 75 | 87 | 0 | 94 |
| | F2,F7,F8,F3,F9,F10,F11,F4 | 75 | 95 | 5 | 94 | 0 | 0 | 18 | 87 | 76 | 89 | 0 | 98 |
| | F1-F11 | 76 | 99 | 5 | 94 | 0 | 87 | 23 | 87 | 76 | 90 | 0 | 96 |
| 10 | F1,F5,F6 | 79 | 89 | 40 | 98 | 0 | 87 | 22 | 97 | 76 | 74 | 16 | 95 |
| | F2,F7,F8 | 74 | 72 | 36 | 98 | 64 | 0 | 0 | 98 | 73 | 59 | 19 | 97 |
| | F3,F9,F10,F11 | 78 | 87 | 36 | 98 | 64 | 0 | 0 | 97 | 77 | 84 | 19 | 97 |
| | F1,F5,F6,F4 | 78 | 89 | 39 | 92 | 0 | 87 | 23 | 87 | 76 | 82 | 22 | 95 |
| | F2,F7,F8,F4 | 74 | 72 | 34 | 92 | 0 | 0 | 18 | 87 | 72 | 58 | 16 | 82 |
| | F3,F9,F10,F11,F4 | 78 | 87 | 36 | 93 | 0 | 0 | 18 | 87 | 78 | 87 | 26 | 97 |
| | F1,F5,F6,F2,F7,F8 | 80 | 96 | 37 | 98 | 0 | 87 | 22 | 97 | 77 | 79 | 20 | 97 |
| | F1,F5,F6,F3,F9,F10,F11 | 79 | 90 | 39 | 98 | 0 | 87 | 22 | 98 | 77 | 84 | 20 | 76 |
| | F2,F7,F8,F3,F9,F10,F11 | 80 | 94 | 37 | 98 | 64 | 0 | 0 | 97 | 78 | 87 | 22 | 80 |
| | F1,F5,F6,F2,F7,F8,F4 | 80 | 96 | 36 | 94 | 0 | 87 | 23 | 87 | 77 | 79 | 22 | 96 |
| | F1,F5,F6,F3,F9,F10,F11,F4 | 79 | 89 | 38 | 94 | 0 | 87 | 23 | 87 | 77 | 84 | 25 | 97 |
| | F2,F7,F8,F3,F9,F10,F11,F4 | 80 | 94 | 36 | 94 | 0 | 0 | 18 | 87 | 78 | 86 | 26 | 98 |
| | F1-F11 | 81 | 99 | 37 | 95 | 0 | 87 | 23 | 87 | 77 | 82 | 23 | 76 |
| 30 | F1,F5,F6 | 83 | 88 | 44 | 98 | 0 | 87 | 22 | 97 | 77 | 80 | 24 | 83 |
| | F2,F7,F8 | 78 | 72 | 41 | 98 | 64 | 0 | 0 | 98 | 69 | 30 | 2 | 96 |
| | F3,F9,F10,F11 | 82 | 87 | 42 | 98 | 64 | 0 | 0 | 97 | 77 | 69 | 18 | 82 |
| | F1,F5,F6,F4 | 83 | 88 | 44 | 93 | 0 | 87 | 23 | 87 | 76 | 76 | 25 | 83 |
| | F2,F7,F8,F4 | 78 | 72 | 40 | 93 | 0 | 0 | 18 | 87 | 73 | 57 | 17 | 83 |
| | F3,F9,F10,F11,F4 | 82 | 87 | 43 | 95 | 0 | 0 | 18 | 87 | 79 | 86 | 32 | 88 |
| | F1,F5,F6,F2,F7,F8 | 84 | 96 | 44 | 98 | 0 | 87 | 22 | 97 | 77 | 74 | 23 | 82 |
| | F1,F5,F6,F3,F9,F10,F11 | 83 | 89 | 44 | 98 | 0 | 87 | 22 | 98 | 75 | 59 | 15 | 84 |
| | F2,F7,F8,F3,F9,F10,F11 | 84 | 94 | 43 | 98 | 64 | 0 | 0 | 97 | 79 | 88 | 28 | 84 |
| | F1,F5,F6,F2,F7,F8,F4 | 84 | 96 | 43 | 96 | 0 | 87 | 23 | 87 | 76 | 68 | 23 | 83 |
| | F1,F5,F6,F3,F9,F10,F11,F4 | 83 | 89 | 44 | 96 | 0 | 87 | 23 | 87 | 79 | 84 | 28 | 84 |
| | F2,F7,F8,F3,F9,F10,F11,F4 | 84 | 94 | 42 | 96 | 0 | 0 | 18 | 87 | 74 | 50 | 15 | 81 |
| | F1-F11 | 85 | 98 | 45 | 97 | 0 | 87 | 23 | 87 | 78 | 86 | 30 | 78 |

, ,  Fuhl and Kasneci

Table 3. For the decision trees, we evaluate the boosting and bagging training algorithms. The input data are the normalized features specified on the left side of the table. Here we evaluated a window length of 41 and used the classifiers in an **offline fashion** meaning that previouse as well as consecutive data points are used in the window for classification of the current data point. *F=Fixation,S=Saccade,P=Smooth Pursuit,E=Error*

| Unit: Features | Bagged RF | | | | Boosted RF | | | | Neural Network | | | |
|---|---|---|---|---|---|---|---|---|---|---|---|---|
| | \multicolumn{12}{c}{*mean Intersection over Union (mIoU) \* 100*} |
| | F | S | P | E | F | S | P | E | F | S | P | E |
| F1,F5,F6 | 92 | 89 | 79 | 98 | 0 | 87 | 22 | 97 | 78 | 73 | 27 | 80 |
| F2,F7,F8 | 84 | 73 | 66 | 98 | 64 | 0 | 0 | 98 | 75 | 61 | 30 | 80 |
| F3,F9,F10,F11 | 89 | 87 | 70 | 98 | 64 | 0 | 0 | 97 | 79 | 78 | 29 | 84 |
| F1,F5,F6,F4 | 91 | 88 | 75 | 94 | 0 | 87 | 23 | 87 | 74 | 48 | 23 | 79 |
| F2,F7,F8,F4 | 83 | 73 | 63 | 93 | 0 | 0 | 18 | 87 | 73 | 48 | 25 | 81 |
| F3,F9,F10,F11,F4 | 89 | 87 | 70 | 95 | 0 | 0 | 18 | 86 | 79 | 74 | 33 | 82 |
| F1,F5,F6,F2,F7,F8 | 91 | 95 | 72 | 98 | 0 | 87 | 22 | 97 | 72 | 27 | 10 | 96 |
| F1,F5,F6,F3,F9,F10,F11 | 92 | 89 | 78 | 98 | 0 | 87 | 22 | 98 | 80 | 84 | 32 | 83 |
| F2,F7,F8,F3,F9,F10,F11 | 91 | 93 | 72 | 98 | 64 | 0 | 0 | 97 | 79 | 74 | 30 | 83 |
| F1,F5,F6,F2,F7,F8,F4 | 91 | 95 | 71 | 95 | 0 | 87 | 23 | 87 | 73 | 38 | 22 | 79 |
| F1,F5,F6,F3,F9,F10,F11,F4 | 92 | 89 | 78 | 96 | 0 | 87 | 23 | 87 | 77 | 59 | 29 | 82 |
| F2,F7,F8,F3,F9,F10,F11,F4 | 91 | 93 | 72 | 96 | 0 | 0 | 18 | 86 | 75 | 46 | 26 | 78 |
| F1-F11 | 93 | 97 | 75 | 97 | 0 | 87 | 23 | 87 | 77 | 61 | 30 | 79 |

The training and test split was chosen so that person one to ten (P1-P10) who drove in the simulator serve as training data and the real drives (P11-P19) as test data. This split was chosen because the data differ significantly (see Figure 3) and it makes the classification much more difficult. Also, we see this scenario as a realistic implementation for generating training data in an industrial setting. Here, the training data would also be recorded in a simulator or on a test track to obtain as much and clean data as possible. As a metric for all evaluations, we used the mean intersection over union (mIoU) or Jaccard index separately for each eye movement as well as the errors ($\frac{Predicted \cap Truth}{Predicted \cup Truth}$). The mIoU was used since it is very sensitive to mistakes. In image based segmentations a score above 0,5 is usually seen as a good result.

Table 2 and Table 3 show the evaluations of different machine learning methods with different feature combinations. In all cases, bagging in combination with RF is the best procedure. For the offline case (Window with previous and future values in Table 3) the results of bagged RF are very good. In the case of a window size of 1 (Table 2) it is very hard to detect the smooth pursuits for all machine learning methods. However, this improves for larger windows. Looking at the individual features (First three evaluations in Table 2 and 3) we can derive the quality ranking F1 best, F3 slightly worse, and F2 worst (F1=pupil center, F2=eye corner vector,F3=optical vector). As for the eye opening degree (F4), it worsens the results in most cases. As for the feature combinations, F1,F5,F6,F2,F7, and F8 seem to be the best in terms of smooth pursuits. If more emphasis is placed on the saccades, F1,F5,F6,F2,F7, and F8 are the best. This can be seen clearly in Table2 and Table3.

## 6 CONCLUSION

In this work, we have presented a new dataset which includes several features and consists of two different recording scenarios. This would be on the one hand the simulator driving and on the other hand the real driving. We have clearly shown that the sequences of eye movements differ significantly (Figure 3) and also performed a feature analysis. The feature analysis can be used to infer which features as well as which combinations can be used effectively. To our knowledge, the presented dataset is currently the largest dataset worldwide.

# REFERENCES

[1] Richard Andersson, Linnea Larsson, Kenneth Holmqvist, Martin Stridh, and Marcus Nyström. 2017. One algorithm to rule them all? An evaluation and discussion of ten eye movement event-detection algorithms. *Behavior Research Methods* 49, 2 (2017), 616–637.

[2] H. Bahmani, W. Fuhl, E. Gutierrez, G. Kasneci, E. Kasneci, and S. Wahl. 2016. Feature-based attentional influences on the accommodation response. In *Vision Sciences Society Annual Meeting Abstract*.

[3] Ravi Teja Chadalavada, Henrik Andreasson, Maike Schindler, Rainer Palm, and Achim J Lilienthal. 2020. Bi-directional navigation intent communication using spatial augmented reality and eye-tracking glasses for improved safety in human–robot interaction. *Robotics and Computer-Integrated Manufacturing* 61 (2020), 101830.

[4] KR Chandrika, J Amudha, and Sithu D Sudarsan. 2019. Identification and Classification of Expertise Using Eye GazeâĂŤIndustrial Use Case Study with Software Engineers. In *International Conference on Communication and Intelligent Systems*. Springer, 391–405.

[5] Leandro L Di Stasi, Michael B McCamy, Stephen L Macknik, James A Mankin, Nicole Hooft, Andrés Catena, and Susana Martinez-Conde. 2014. Saccadic eye movement metrics reflect surgical residents' fatigue. *Annals of surgery* 259, 4 (2014), 824–829.

[6] Andrew T Duchowski and Sophie Jörg. 2015. Modeling Physiologically Plausible Eye Rotations. *Proceedings of Computer Graphics International* (2015).

[7] S Eivazi, W Fuhl, and E Kasneci. 2017. Towards intelligent surgical microscopes: Surgeons gaze and instrument tracking. In *Proceedings of the 22st International Conference on Intelligent User Interfaces, IUI*.

[8] Shahram Eivazi, Michael Slupina, Wolfgang Fuhl, Hoorieh Afkari, Ahmad Hafez, and Enkelejda Kasneci. 2017. Towards automatic skill evaluation in microsurgery. In *Proceedings of the 22st International Conference on Intelligent User Interfaces, IUI 2017*. ACM.

[9] Ralf Engbert and Reinhold Kliegl. 2003. Microsaccades uncover the orientation of covert attention. *Vision research* 43, 9 (2003), 1035–1045.

[10] W. Fuhl. 2019. *Image-based extraction of eye features for robust eye tracking*. Ph.D. Dissertation. University of TÃijbingen.

[11] Wolfgang Fuhl. 2020. From perception to action using observed actions to learn gestures. *User Modeling and User-Adapted Interaction* (08 2020), 1–18.

[12] Wolfgang Fuhl, Efe Bozkir, Benedikt Hosp, Nora Castner, David Geisler, Thiago C., and Enkelejda Kasneci. 2019. Encodji: Encoding Gaze Data Into Emoji Space for an Amusing Scanpath Classification Approach ;). In *Eye Tracking Research and Applications*.

[13] Wolfgang Fuhl, Efe Bozkir, and Enkelejda Kasneci. 2020. Reinforcement learning for the privacy preservation and manipulation of eye tracking data. *arXiv preprint arXiv:2002.06806* (08 2020).

[14] W. Fuhl, N. Castner, and E. Kasneci. 2018. Histogram of oriented velocities for eye movement detection. In *International Conference on Multimodal Interaction Workshops, ICMIW*.

[15] Wolfgang Fuhl, Nora Castner, and Enkelejda Kasneci. 2018. Histogram of oriented velocities for eye movement detection. In *Proceedings of the Workshop on Modeling Cognitive Processes from Multimodal Data*. ACM, 5.

[16] W. Fuhl, N. Castner, and E. Kasneci. 2018. Rule based learning for eye movement type detection. In *International Conference on Multimodal Interaction Workshops, ICMIW*.

[17] Wolfgang Fuhl, Nora Castner, and Enkelejda Kasneci. 2018. Rule-based learning for eye movement type detection. In *Proceedings of the Workshop on Modeling Cognitive Processes from Multimodal Data*. ACM, 9.

[18] W. Fuhl, N. Castner, T. C. KÃijbler, A. Lotz, W. Rosenstiel, and E. Kasneci. 2019. Ferns for area of interest free scanpath classification. In *Proceedings of the 2019 ACM Symposium on Eye Tracking Research & Applications (ETRA)*.

[19] W. Fuhl, N. Castner, L. Zhuang, M. Holzer, W. Rosenstiel, and E. Kasneci. 2018. MAM: Transfer learning for fully automatic video annotation and specialized detector creation. In *International Conference on Computer Vision Workshops, ICCVW*.

[20] W. Fuhl, S. Eivazi, B. Hosp, A. Eivazi, W. Rosenstiel, and E. Kasneci. 2018. BORE: Boosted-oriented edge optimization for robust, real time remote pupil center detection. In *Eye Tracking Research and Applications, ETRA*.

[21] W. Fuhl, H. Gao, and E. Kasneci. 2020. Neural networks for optical vector and eye ball parameter estimation. In *ACM Symposium on Eye Tracking Research & Applications, ETRA 2020*. ACM.

[22] W. Fuhl, D. Geisler, W. Rosenstiel, and E. Kasneci. 2019. The applicability of Cycle GANs for pupil and eyelid segmentation, data generation and image refinement. In *International Conference on Computer Vision Workshops, ICCVW*.

[23] W. Fuhl, D. Geisler, T. Santini, T. Appel, W. Rosenstiel, and E. Kasneci. 2018. CBF:Circular binary features for robust and real-time pupil center detection. In *ACM Symposium on Eye Tracking Research & Applications*.

[24] W. Fuhl, D. Geisler, T. Santini, and E. Kasneci. 2016. Evaluation of State-of-the-Art Pupil Detection Algorithms on Remote Eye Images. In *ACM International Joint Conference on Pervasive and Ubiquitous Computing: Adjunct publication – PETMEI 2016*.

[25] W. Fuhl and E. Kasneci. 2018. Eye movement velocity and gaze data generator for evaluation, robustness testing and assess of eye tracking software and visualization tools. In *Poster at Egocentric Perception, Interaction and Computing, EPIC*.

[26] Wolfgang Fuhl and Enkelejda Kasneci. 2018. Eye movement velocity and gaze data generator for evaluation, robustness testing and assess of eye tracking software and visualization tools. *CoRR* abs/1808.09296 (2018). arXiv:1808.09296 http://arxiv.org/abs/1808.09296

[27] W. Fuhl and E. Kasneci. 2019. Learning to validate the quality of detected landmarks. In *International Conference on Machine Vision, ICMV*.

[28] Wolfgang Fuhl and Enkelejda Kasneci. 2020. Multi Layer Neural Networks as Replacement for Pooling Operations. *arXiv preprint arXiv:2006.06969* (08 2020).

[29] Wolfgang Fuhl and Enkelejda Kasneci. 2020. Rotated Ring, Radial and Depth Wise Separable Radial Convolutions. *arXiv* (08 2020).

[30] Wolfgang Fuhl and Enkelejda Kasneci. 2020. Weight and Gradient Centralization in Deep Neural Networks. *arXiv* (08 2020).

[31] W. Fuhl, G. Kasneci, W. Rosenstiel, and E. Kasneci. 2020. Training Decision Trees as Replacement for Convolution Layers. In *Conference on Artificial Intelligence, AAAI*.

[32] W. Fuhl, T. KÃijbler, T. Santini, and E. Kasneci. 2018. Automatic generation of saliency-based areas of interest. In *Symposium on Vision, Modeling and Visualization (VMV)*.

[33] W. Fuhl, T. C. KÃijbler, H. Brinkmann, R. Rosenberg, W. Rosenstiel, and E. Kasneci. 2018. Region of interest generation algorithms for eye tracking data. In *Third Workshop on Eye Tracking and Visualization (ETVIS), in conjunction with ACM ETRA*.

[34] W. Fuhl, T. C. KÃijbler, D. Hospach, O. Bringmann, W. Rosenstiel, and E. Kasneci. 2017. Ways of improving the precision of eye tracking data: Controlling the influence of dirt and dust on pupil detection. *Journal of Eye Movement Research* 10, 3 (05 2017).

[35] W. Fuhl, T. C. KÃijbler, K. Sippel, W. Rosenstiel, and E. Kasneci. 2015. Arbitrarily shaped areas of interest based on gaze density gradient. In *European Conference on Eye Movements, ECEM 2015*.

[36] W. Fuhl, T. C. KÃijbler, K. Sippel, W. Rosenstiel, and E. Kasneci. 2015. ExCuSe: Robust Pupil Detection in Real-World Scenarios. In *16th International Conference on Computer Analysis of Images and Patterns (CAIP 2015)*.

[37] Wolfgang Fuhl, Yao Rong, and Kasneci Enkelejda. 2020. Fully Convolutional Neural Networks for Raw Eye Tracking Data Segmentation, Generation, and Reconstruction. In *Proceedings of the International Conference on Pattern Recognition*. 0–0.

[38] Wolfgang Fuhl, Yao Rong, Thomas Motz, Michael Scheidt, Andreas Hartel, Andreas Koch, and Enkelejda Kasneci. 2020. Explainable Online Validation of Machine Learning Models for Practical Applications. *arXiv* (08 2020).

[39] W. Fuhl, W. Rosenstiel, and E. Kasneci. 2019. 500,000 images closer to eyelid and pupil segmentation. In *Computer Analysis of Images and Patterns, CAIP*.

[40] W. Fuhl, T. Santini, D. Geisler, T. C. KÃijbler, and E. Kasneci. 2017. EyeLad: Remote Eye Tracking Image Labeling Tool. In *12th Joint Conference on Computer Vision, Imaging and Computer Graphics Theory and Applications (VISIGRAPP 2017)*.

[41] W. Fuhl, T. Santini, D. Geisler, T. C. KÃijbler, W. Rosenstiel, and E. Kasneci. 2016. Eyes Wide Open? Eyelid Location and Eye Aperture Estimation for Pervasive Eye Tracking in Real-World Scenarios. In *ACM International Joint Conference on Pervasive and Ubiquitous Computing: Adjunct publication – PETMEI 2016*.

[42] W. Fuhl, T. Santini, and E. Kasneci. 2017. Fast and Robust Eyelid Outline and Aperture Detection in Real-World Scenarios. In *IEEE Winter Conference on Applications of Computer Vision (WACV 2017)*.

[43] W. Fuhl, T. Santini, and E. Kasneci. 2017. Fast camera focus estimation for gaze-based focus control. In *CoRR*.

[44] W. Fuhl, T. Santini, G. Kasneci, and E. Kasneci. 2016. PupilNet: Convolutional Neural Networks for Robust Pupil Detection. In *CoRR*.

[45] W. Fuhl, T. Santini, G. Kasneci, and E. Kasneci. 2017. PupilNet v2.0: Convolutional Neural Networks for Robust Pupil Detection. In *CoRR*.

[46] W. Fuhl, T. Santini, T. C. KÃijbler, and E. Kasneci. 2016. ElSe: Ellipse Selection for Robust Pupil Detection in Real-World Environments. In *Proceedings of the Ninth Biennial ACM Symposium on Eye Tracking Research & Applications (ETRA)*. 123–130.

[47] W. Fuhl, T. Santini, T. Kuebler, N. Castner, W. Rosenstiel, and E. Kasneci. 2018. Eye movement simulation and detector creation to reduce laborious parameter adjustments. *arXiv preprint arXiv:1804.00970* (2018).

[48] Wolfgang Fuhl, Thiago Santini, Thomas Kuebler, Nora Castner, Wolfgang Rosenstiel, and Enkelejda Kasneci. 2018. Eye movement simulation and detector creation to reduce laborious parameter adjustments. *eprint arXiv:1804.00970* (2018).

[49] W. Fuhl, T. Santini, C. Reichert, D. Claus, A. Herkommer, H. Bahmani, K. Rifai, S. Wahl, and E. Kasneci. 2016. Non-Intrusive Practitioner Pupil Detection for Unmodified Microscope Oculars. *Elsevier Computers in Biology and Medicine*

79 (12 2016), 36–44.
[50] Wolfgang Fuhl, Marc Tonsen, Andreas Bulling, and Enkelejda Kasneci. 2016. Pupil detection for head-mounted eye tracking in the wild: An evaluation of the state of the art. In *Machine Vision and Applications*. 1–14.
[51] D. Geisler, W. Fuhl, T. Santini, and E. Kasneci. 2017. Saliency Sandbox: Bottom-Up Saliency Framework. In *12th Joint Conference on Computer Vision, Imaging and Computer Graphics Theory and Applications (VISIGRAPP 2017)*.
[52] Kenneth Holmqvist, Marcus Nyström, Richard Andersson, Richard Dewhurst, Halszka Jarodzka, and Joost Van de Weijer. 2011. *Eye tracking: A comprehensive guide to methods and measures*. OUP Oxford.
[53] Sabrina Hoppe and Andreas Bulling. 2016. End-to-end eye movement detection using convolutional neural networks. *arXiv preprint arXiv:1609.02452* (2016).
[54] Qiang Ji, Zhiwei Zhu, and Peilin Lan. 2004. Real-time nonintrusive monitoring and prediction of driver fatigue. *IEEE transactions on vehicular technology* 53, 4 (2004), 1052–1068.
[55] T. C. Kübler, K. Sippel, W. Fuhl, G. Schievelbein, J. Aufreiter, R. Rosenberg, W. Rosenstiel, and E. Kasneci. 2015. *Analysis of eye movements with Eyetrace*. Vol. 574. Biomedical Engineering Systems and Technologies. Communications in Computer and Information Science (CCIS). Springer International Publishing. 458–471 pages.
[56] Petr Kellnhofer, Adria Recasens, Simon Stent, Wojciech Matusik, and Antonio Torralba. 2019. Gaze360: Physically unconstrained gaze estimation in the wild. In *Proceedings of the IEEE International Conference on Computer Vision*. 6912–6921.
[57] Oleg V Komogortsev, Denise V Gobert, Sampath Jayarathna, Do Hyong Koh, and Sandeep M Gowda. 2010. Standardization of automated analyses of oculomotor fixation and saccadic behaviors. *IEEE Transactions on Biomedical Engineering* 57, 11 (2010), 2635–2645.
[58] Oleg V Komogortsev and Javed I Khan. 2009. Eye movement prediction by oculomotor plant Kalman filter with brainstem control. *Journal of Control Theory and Applications* 7, 1 (2009), 14–22.
[59] Rakshit Kothari, Zhizhuo Yang, Christopher Kanan, Reynold Bailey, Jeff B Pelz, and Gabriel J Diaz. 2020. Gaze-in-wild: A dataset for studying eye and head coordination in everyday activities. *Scientific reports* 10, 1 (2020), 1–18.
[60] Rakshit S Kothari, Aayush K Chaudhary, Reynold J Bailey, Jeff B Pelz, and Gabriel J Diaz. 2020. EllSeg: An Ellipse Segmentation Framework for Robust Gaze Tracking. *arXiv preprint arXiv:2007.09600* (2020).
[61] Matthias Kummerer, Thomas SA Wallis, Leon A Gatys, and Matthias Bethge. 2017. Understanding low-and high-level contributions to fixation prediction. In *Proceedings of the IEEE International Conference on Computer Vision*. 4789–4798.
[62] Linnéa Larsson, Marcus Nyström, Richard Andersson, and Martin Stridh. 2015. Detection of fixations and smooth pursuit movements in high-speed eye-tracking data. *Biomedical Signal Processing and Control* 18 (2015), 145–152.
[63] Linnéa Larsson, Marcus Nyström, and Martin Stridh. 2013. Detection of saccades and postsaccadic oscillations in the presence of smooth pursuit. *IEEE Transactions on Biomedical Engineering* 60, 9 (2013), 2484–2493.
[64] Sandra P Marshall. 2007. Identifying cognitive state from eye metrics. *Aviation, space, and environmental medicine* 78, 5 (2007), B165–B175.
[65] James G May, Robert S Kennedy, Mary C Williams, William P Dunlap, and Julie R Brannan. 1990. Eye movement indices of mental workload. *Acta psychologica* 75, 1 (1990), 75–89.
[66] Martin Fodslette Møller. 1993. A scaled conjugate gradient algorithm for fast supervised learning. *Neural networks* 6, 4 (1993), 525–533.
[67] Marcus Nyström and Kenneth Holmqvist. 2010. An adaptive algorithm for fixation, saccade, and glissade detection in eyetracking data. *Behavior research methods* 42, 1 (2010), 188–204.
[68] Anantha M Prasad, Louis R Iverson, and Andy Liaw. 2006. Newer classification and regression tree techniques: bagging and random forests for ecological prediction. *Ecosystems* 9, 2 (2006), 181–199.
[69] Dario D Salvucci and Joseph H Goldberg. 2000. Identifying fixations and saccades in eye-tracking protocols. In *Eye Tracking Research and Applications*. ACM, 71–78.
[70] Chris Seiffert, Taghi M Khoshgoftaar, Jason Van Hulse, and Amri Napolitano. 2009. RUSBoost: A hybrid approach to alleviating class imbalance. *IEEE Transactions on Systems, Man, and Cybernetics-Part A: Systems and Humans* 40, 1 (2009), 185–197.
[71] Oleg Špakov, Diederick Niehorster, Howell Istance, Kari-Jouko Räihä, and Harri Siirtola. 2019. Two-Way Gaze Sharing in Remote Teaching. In *IFIP Conference on Human-Computer Interaction*. Springer, 242–251.
[72] Enkelejda Tafaj, Gjergji Kasneci, Wolfgang Rosenstiel, and Martin Bogdan. 2012. Bayesian online clustering of eye movement data. In *Proceedings of the symposium on eye tracking research and applications*. 285–288.
[73] Ralf van der Lans, Michel Wedel, and Rik Pieters. 2011. Defining eye-fixation sequences across individuals and tasks: the Binocular-Individual Threshold (BIT) algorithm. *Behavior Research Methods* 43, 1 (2011), 239–257.
[74] Giacomo Veneri, Pietro Piu, Pamela Federighi, Francesca Rosini, Antonio Federico, and Alessandra Rufa. 2010. Eye fixations identification based on statistical analysis-case study. In *Cognitive Information Processing (CIP), 2010 2nd International Workshop on*. IEEE, 446–451.

[75] Giacomo Veneri, Pietro Piu, Francesca Rosini, Pamela Federighi, Antonio Federico, and Alessandra Rufa. 2011. Automatic eye fixations identification based on analysis of variance and covariance. *Pattern Recognition Letters* 32, 13 (2011), 1588–1593.
[76] Raimondas Zemblys, Diederick C Niehorster, and Kenneth Holmqvist. 2019. gazeNet: End-to-end eye-movement event detection with deep neural networks. *Behavior research methods* 51, 2 (2019), 840–864.
[77] Raimondas Zemblys, Diederick C Niehorster, Oleg Komogortsev, and Kenneth Holmqvist. 2018. Using machine learning to detect events in eye-tracking data. *Behavior research methods* 50, 1 (2018), 160–181.